\documentclass[11pt, a4paper]{article}

\usepackage{geometry}
\usepackage{authblk} 

\usepackage[T1]{fontenc}
\usepackage{multirow}
\usepackage{graphicx}
\usepackage{pgfplots}
\usepackage{tikz}
\usetikzlibrary{arrows.meta, positioning, shapes.multipart, fit}
\usepackage{amsmath}
\usepackage{amssymb}
\usepackage{amsfonts}
\usepackage{enumitem}
\usepackage{booktabs}
\usepackage{wrapfig}
\usepackage{float}
\usepackage{makecell} 
\usepackage{tabularx}
\usepackage{orcidlink}
\usepackage{xcolor}
\usepackage{bbding}
\usepackage{cite}
\usepackage{eso-pic}

\usepackage{hyperref}
\hypersetup{
    colorlinks=true,
    linkcolor=blue,
    filecolor=magenta,      
    urlcolor=cyan,
    citecolor=blue,
}

\definecolor{softblue}{RGB}{230,240,255}
\definecolor{softred}{RGB}{255,235,235}
\definecolor{softgreen}{RGB}{230,250,235}
\tikzset{
  box/.style = {rounded corners, draw=black, fill=white, very thick, align=left, inner sep=6pt},
  title/.style = {font=\bfseries},
  bad/.style = {box, fill=softred},
  good/.style = {box, fill=softgreen},
  hub/.style  = {rounded corners, draw=black, fill=softblue, very thick, align=center, inner sep=6pt},
  arrow/.style = {-{Latex[round]}, thick},
  dashedarrow/.style = {dash pattern=on 2pt off 1.5pt, -{Latex[round]}, thick}
}
\pgfplotsset{compat=1.17}

\newcolumntype{Y}{>{\centering\arraybackslash}X}
\setlist[itemize]{label=\textbullet}
\setlist[enumerate]{itemsep=0mm}

\title{SentiFuse: Deep Multi-model Fusion Framework for Robust Sentiment Extraction}

\author[1]{Hieu Minh Duong\orcidlink{0000-0002-4715-8169}\thanks{Corresponding author: \texttt{hieu.duong@louisville.edu}}}
\author[1]{Rupa Ghosh\orcidlink{0009-0004-1550-8314}}
\author[1]{Cong Hoan Nguyen\orcidlink{0009-0007-5046-1453}}
\author[2]{Eugene Levin\orcidlink{0000-0002-8127-3208}}
\author[2]{Todd Gary\orcidlink{0000-0003-4121-6628}}
\author[1]{Long Nguyen\orcidlink{0000-0001-7673-7955}}

\affil[1]{University of Louisville, Louisville, Kentucky, USA \protect\\ \texttt{\{hieu.duong,rupa.ghosh,conghoan.nguyen,l.nguyen\}@louisville.edu}}
\affil[2]{Meharry Medical College, Nashville, Tennessee, USA \protect\\ \texttt{\{tpgary,elevin\}@mmc.edu}}

\date{} 

\begin{document}
\AddToShipoutPictureBG*{%
  \AtPageLowerLeft{%
    \put(\LenToUnit{0.5\paperwidth},\LenToUnit{1.5cm}){%
      \makebox[0pt][c]{%
        \begin{minipage}{0.9\paperwidth}
          \centering \small
          Accepted at the 14th International Conference on Computational Data and Social Networks (CSoNet 2025).
        \end{minipage}%
      }
    }
  }
}
\maketitle

\begin{abstract}
Sentiment analysis models exhibit complementary strengths, yet existing approaches lack a unified framework for effective integration. We present SentiFuse, a flexible and model-agnostic framework that integrates heterogeneous sentiment models through a standardization layer and multiple fusion strategies. Our approach supports decision-level fusion, feature-level fusion, and adaptive fusion, enabling systematic combination of diverse models. We conduct experiments on three large-scale social-media datasets: Crowdflower, GoEmotions, and Sentiment140. These experiments show that SentiFuse consistently outperforms individual models and naive ensembles. Feature-level fusion achieves the strongest overall effectiveness, yielding up to 4\% absolute improvement in F1 score over the best individual model and simple averaging, while adaptive fusion enhances robustness on challenging cases such as negation, mixed emotions, and complex sentiment expressions. These results demonstrate that systematically leveraging model complementarity yields more accurate and reliable sentiment analysis across diverse datasets and text types.

\vspace{0.5cm}
\end{abstract}

\section{Introduction}

Large Language Models (LLMs) and deep learning architectures have transformed sentiment analysis by capturing semantic dependencies and contextual nuances beyond the reach of traditional approaches. Rule-based systems such as VADER \cite{Hutto2014} offer interpretability and efficiency but are brittle in the presence of sarcasm, domain-specific language, or complex modifiers. Statistical models provide lightweight computation but cannot adequately capture sequential dependencies and contextual relationships \cite{10.1145/2436256.2436274}. In contrast, deep learning models excel in modeling semantics but demand large labeled datasets and risk overfitting \cite{REUSENS2024124302}. These complementary strengths and weaknesses are especially evident in social media, where sentiment expressions are highly contextual and often include irony, mixed emotions, or pragmatic cues that challenge single-model methods \cite{RODRIGUEZIBANEZ2023119862,RAVI201514}. For instance, the sentence “Great, another delayed flight – exactly what I needed today!” is likely to be misclassified by a lexicon-based model as positive due to the words great and needed, whereas a context-sensitive model could recognize the sarcastic intent. Such cases highlight both the limitations of individual sentiment analysis approaches and the potential gains from systematically integrating heterogeneous models to leverage their complementary perspectives.

Despite the promise of ensemble learning, existing sentiment analysis fusion methods are typically simplistic and ad hoc. Majority voting, unweighted averaging, and confidence-weighted schemes assume homogeneous base models with compatible outputs and aligned decision boundaries \cite{mohammad2013nrc,kiritchenko2014nrc}. These naive methods neglect fundamental challenges in heterogeneous model integration, such as inconsistencies in output formats (probabilities, logits, discrete labels), disparities in confidence calibration, and the lack of adaptive mechanisms that adjust to varying linguistic complexities. To address these gaps, we propose SentiFuse, a model-agnostic framework that integrates heterogeneous sentiment models through standardized output processing and multiple fusion strategies. Our work focuses on three Research Questions: \textit{\textbf{Does systematic fusion outperform naive combination methods?} }(RQ1), \textit{\textbf{How does fusion effectiveness vary across text characteristics?}} (RQ2), \textit{\textbf{Does the framework generalize across different model combinations?}} (RQ3). To this end, our contributions are threefold.
\begin{itemize}
    \item First, we introduce a standardization layer that unifies diverse outputs into probability distributions, enabling integration without architectural modifications.
    \item Second, we design three complementary fusion strategies—decision-level fusion with learned weights, feature-level fusion using meta-classification, and adaptive fusion guided by automatically extracted text characteristics.
    \item Finally, we conduct a comprehensive evaluation based on three Research Questions, demonstrating that SentiFuse improves performance over both standalone models and traditional ensembles across diverse and complex sentiment scenarios.
\end{itemize}

\section{Related Work}
Early sentiment analysis relied on statistical methods such as TF-IDF vectorization with traditional classifiers \cite{SALTON1988513,Dey2023,Das2018,chikersal-etal-2015-sentu}, which were efficient but ignored sequential dependencies and semantics. Rule-based systems like VADER \cite{Hutto2014} and SentiWordNet \cite{baccianella2010sentiwordnet} improved interpretability with linguistic heuristics, yet struggled with context sensitivity and domain adaptation \cite{elbagir2019twitter,Ray2022,taboada2011lexicon,Chiny2021}. Deep learning models \cite{10.1007/s10462-019-09794-5,wang2023multilevel,app12083709,Behera2021} addressed these issues: recurrent architectures such as BiLSTM captured long-range dependencies \cite{xu2019bilstm}, while attention mechanisms highlighted sentiment-bearing spans \cite{zhou2016attention}. Transformer-based models, notably BERT \cite{devlin2018bert} and RoBERTa \cite{liu2019roberta}, achieved state-of-the-art results with bidirectional context encoding, though at the cost of data demands and sensitivity to domain shifts.

Beyond single models, ensemble methods have been widely applied. Traditional approaches like bagging \cite{breiman1996bagging} and voting \cite{dietterich2000ensemble} improved robustness but failed to fully exploit heterogeneous systems. More recent meta-ensembles \cite{kora2023meta} and adaptive fusion strategies \cite{gan2024multimodal} dynamically weight model outputs or modalities, yielding stronger results but often requiring high computational resources or paired multimodal data. Meanwhile, the emergence of large language models (LLMs) has shifted sentiment analysis research: studies report ChatGPT achieving competitive or superior performance in zero- and few-shot scenarios \cite{wang2024chatgpt} compared to fine-tuned transformers \cite{lossio2024comparison}. Cross-lingual ensembles also demonstrate strong results by combining translation with transformer ensembles \cite{miah2024ensemble}, though error propagation remains a challenge. Recent surveys emphasize that many fusion approaches still rely on static weighting and lack systematic evaluation against simpler baselines \cite{singh2024survey}, which underscores the need for model-agnostic, adaptive frameworks.

\section{Methodology}
In this study, we propose an innovative multi-model framework, SentiFuse, designed to integrate multiple heterogeneous sentiment analysis models. Our proposed sentiment analysis framework is composed of four interconnected components: (1) multiple heterogeneous sentiment analysis models, (2) a standardization layer, (3) a fusion strategy selector, and (4) sentiment classification. This structured approach enables flexible integration and coherent combination of sentiment predictions from diverse model types, ranging from simple lexicon-based methods to complex language models.

\begin{figure}[!htbp]
    \centering
    \includegraphics[width=0.6\linewidth]{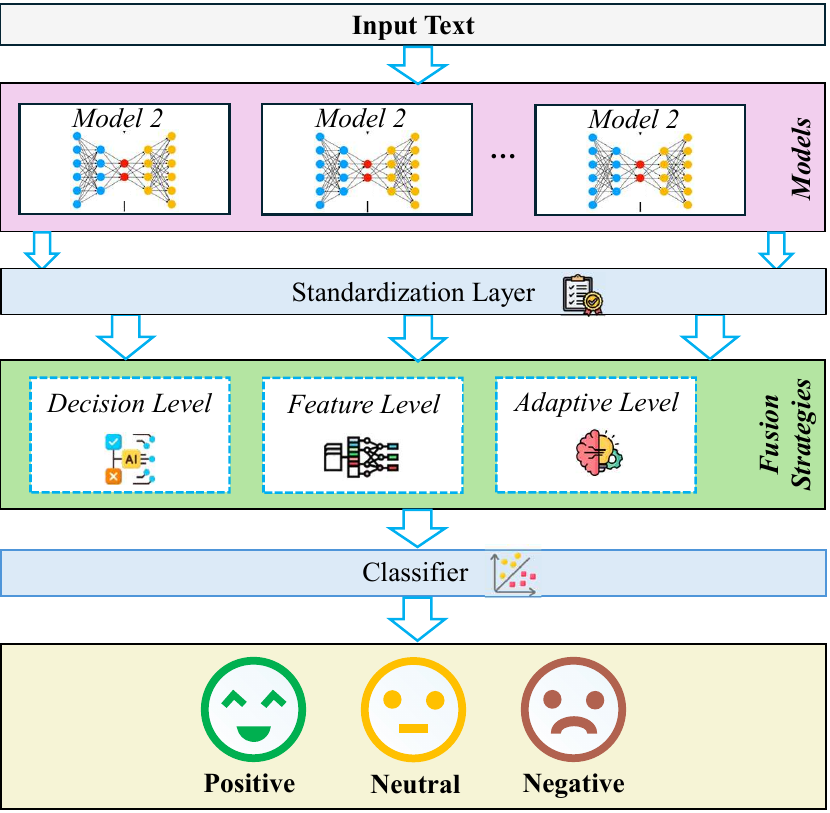}
    \caption{Overall architecture of the proposed SentiFuse framework.}
    \label{fig:Architecture}
\end{figure}

\subsection{Multi-model Sentiment Analysis.}
The first component consists of a diverse set of sentiment models represented as:

\begin{equation}
    M = {M_1, M_2,..., M_n}.
\end{equation}

Each model $M_i$ processes input text to produce an output $O_i = M_i(x)$. Our implementation specifically includes:

\begin{itemize}[leftmargin=10pt, topsep=0.5pt,itemsep=-0.5ex,partopsep=1ex,parsep=1ex]
  \item \textbf{Lexicon-based models:} Estimate sentiment by aggregating word-level polarity from curated lexicons, typically with heuristics for intensity, positional emphasis, and valence shifters (e.g., negation).
  \item \textbf{Pattern-based models:} Infer sentiment from the presence and strength of predefined sentiment-bearing patterns (e.g., idioms, emoji sequences, dependency templates), weighting patterns by frequency and reliability.
  \item \textbf{Machine-learning models:} Learn a mapping from engineered features (e.g., n-grams, TF--IDF, syntactic cues, lexicon features) to sentiment labels using supervised classifiers (e.g., logistic regression, SVM).
  \item \textbf{Encoding models:} Use deep contextual encoders (e.g., BERT, RoBERTa) to obtain sequence representations; a pooled vector (e.g., [CLS] token) is passed to a task-specific classifier after pretraining and fine-tuning.
\end{itemize}
We distinguish classical classifiers trained on engineered features from neural encoding models that learn contextual representations; for classification, the encoder is paired with a task-specific prediction head and fine-tuned.

\subsection{Standardization Layer.}
To facilitate seamless integration of heterogeneous model outputs, we implement a standardization function $S$ that converts different model outputs into unified probability distributions over sentiment classes as:

\begin{equation}
    S\left(O_{i}\right)=\left\{\begin{array}{ll}
\left\{p_{\text {pos }}, p_{\text {neg }}\right\}, & \text { if output is probability} \\
\left\{\frac{1+s}{2}, \frac{1-s}{2}\right\}, & \text { if output is score } s \in[-1,1] \\
\left\{\sigma\left(v_{\text {pos }}\right), \sigma\left(v_{\text {neg }}\right)\right\}, & \text { if output is logits }
\end{array}\right..
\end{equation}

Additionally, we define feature extraction functions $\phi_{i}\left(O_{i}\right)$ specific to each model type as: 
\begin{equation}
    \phi_{i}\left(O_{i}\right)=\left[f_{1}\left(O_{i}\right), f_{2}\left(O_{i}\right), \ldots, f_{k}\left(O_{i}\right)\right]^{\top},
\end{equation}
where each $f_i$ represents a feature extraction operation regarding the model type.

\subsection{Fusion Strategies.}
We formalize three distinct fusion methods:
\begin{itemize}[leftmargin=10pt, topsep=1pt,itemsep=-0.5ex,partopsep=1ex,parsep=1ex]
    \item \textbf{Decision-level Fusion}: Combines standardized probability outputs from each model using weighted averages defined as follows: 
\begin{equation}
    F_{\text {d }}\left(S\left(O_{1}\right), \ldots, S\left(O_{n}\right)\right) = \frac{\sum_{i=1}^{n} w_{i} \cdot S\left(O_{i}\right)}{\sum_{i=1}^{n} w_{i}},
\end{equation}
with model-specific weights $w_i \in [0, 1]$.
    \item \textbf{Feature-level Fusion}: Aggregates extracted features from multiple models into a unified vector for classification represented as follows:
\begin{equation}
    F_{\text {f}}\left(S\left(O_{1}\right), \ldots, S\left(O_{n}\right)\right) = g\left(\left[\phi_{1}\left(O_{1}\right) \oplus \ldots \oplus \phi_{n}\left(O_{n}\right)\right]\right),
\end{equation}
where $\oplus$ indicates concatenation, $\phi_i$ is a feature mapping for model $i$, and $g$ is a trained meta-classifier
    \item \textbf{Adaptive Fusion:} Dynamically re-weights model contributions based on text characteristics defined as:
\begin{equation}
F_{\text{a}}\!\left(x, S(O_{1}), \ldots, S(O_{n})\right) =
\frac{\sum_{i=1}^{n} w_{i}(x) \cdot S(O_{i})}{\sum_{i=1}^{n} w_{i}(x)},
\end{equation}
where $w_i(x)$ are adaptive weights determined from textual features $\psi(x)$
(e.g., negation presence, text length, emotional complexity).
\end{itemize}

\subsection{Sentiment Classification.}
The fused output is mapped to sentiment probabilities by a classification function defined as follows:
\begin{equation}
    C\left(F\left(S\left(O_{1}\right), \ldots, S\left(O_{n}\right)\right)\right)=\left\{p_{c_{1}}, p_{c_{2}}, \ldots, p_{c_{k}}\right\},
\end{equation}
where $p_{c_i}$ is the predicted probability for sentiment class $c_i$. The final sentiment label $L(x)$ is determined using a confidence threshold $\delta$. For example, if the dataset has three types of label as follows:
\begin{equation}
    L(x)=\left\{\begin{array}{ll}
\text { positive, } & p_{\text {pos }}>p_{\text {neg }}+\delta \\
\text { negative, } & p_{\text {neg }}>p_{\text {pos }}+\delta \\
\text { neutral, } & \text { otherwise }
\end{array}\right..
\end{equation}
This generalized formulation allows our framework to seamlessly handle multiclass sentiment tasks and provides greater flexibility in diverse sentiment analysis scenarios.

\subsection{Training and Adaptation. }
Fusion weights are trained on labeled data. Decision-level fusion tunes weights on validation sets, while feature-level fusion uses logistic regression with L2 regularization. Adaptive fusion adjusts weights by text type: transformers get more weight with negation or mixed emotions, lexicons with short texts. Models start equal and are modified by simple rules, keeping the system efficient and easy to extend.

\section{Experiments and Evaluation}
\subsection{Experiment Setup}
\subsubsection{Datasets.} We evaluate SentiFuse on three sentiment datasets:
\begin{itemize}[leftmargin=10pt, topsep=3pt,itemsep=-0.5ex,partopsep=1ex,parsep=1ex]
    \item \textbf{Crowdflower US Airline Twitter} (14.6k tweets): A benchmark dataset with balanced sentiment labels (positive, negative, neutral) focusing on airline customer experiences.
    \item \textbf{GoEmotions} (211k Reddit posts) \cite{goemo}: A comprehensive emotion dataset from Google, containing 28 fine-grained emotions categorized into positive, negative, and neutral sentiment classes.
    \item \textbf{Sentiment140} (1.6M tweets) \cite{sent140}: A large-scale tweet dataset for sentiment classification, labeled as negative, neutral, positive.
\end{itemize}

All datasets undergo consistent preprocessing with text normalization and sentiment label standardization. We apply 80-10-10 stratified splits for training, validation, and testing.

\subsubsection{Baseline models.}
To evaluate our framework, we deliberately employ a diverse set of sentiment analysis models that represent different methodological paradigms: 
 \begin{itemize}[leftmargin=10pt, topsep=3pt,itemsep=-0.5ex,partopsep=1ex,parsep=1ex]
     \item \textbf{Classical machine learning models}. We incorporate TF–IDF vectorization combined with linear classifiers such as SVM and XGBoost. These models rely on bag-of-words style representations, which capture lexical patterns effectively but ignore deep contextual semantics.
     \item \textbf{Deep neural models}. To represent state-of-the-art contextual embeddings, we include BERT, RoBERTa, and DistilBERT. The versions used in the experiments are uncased from Hugging Face Transformers, pretrained on English Wikipedia and BookCorpus. No additional pretraining was performed; we fine-tuned these models directly on the sentiment datasets. These transformers encode rich semantic and syntactic information, leading to strong performance across benchmarks but at higher computational cost.

 \end{itemize}
For fair comparison, we also evaluate several standard ensemble rules: simple averaging, confidence-weighted averaging, majority voting, median averaging, and max-confidence selection.
 
\subsection{Research Question 1}
The first research question, "\textit{\textbf{Does systematic fusion outperform naive combination methods?}}", evaluates whether structured fusion strategies provide measurable benefits over naive ensemble rules and individual models. The fixed model pool for this analysis consists of \textbf{VADER}, \textbf{DistilBERT}, and a \textbf{TF--IDF} classifier, chosen to represent lexicon-based, neural, and statistical paradigms. We compare three categories of methods: (i) the best-performing individual model (typically DistilBERT), (ii) naive ensembles including simple averaging, confidence-weighted averaging, majority voting, median averaging, and max-confidence selection, and (iii) structured fusion methods, namely decision-level, feature-level, and adaptive fusion. Table \ref{tab:rq1_results} reports Accuracy, Precision, Recall, F1-Score of strategies across datasets.

\begin{table}[!htbp]
\centering
\caption{Performance (percentage) of fusion strategies across datasets.}
\label{tab:rq1_results}
\resizebox{\textwidth}{!}{%
\begin{tabular}{l|cccc|cccc|cccc}
\toprule

\multirow{2}{*}{Strategy} 
& \multicolumn{4}{c|}{Crowdflower} 
& \multicolumn{4}{c|}{GoEmotions} 
& \multicolumn{4}{c}{Sentiment140} \\
\cmidrule(lr){2-5} \cmidrule(lr){6-9} \cmidrule(lr){10-13}
& Recall & Prec & Acc & F1 
& Recall & Prec & Acc & F1 
& Recall & Prec & Acc & F1 \\
\midrule
Best Individual & 77.80 & 58.79 & 87.60 & 66.97 & 53.59 & \textbf{77.29} & 77.32 & 63.29 & 77.28 & 75.60 & 76.11 & 76.43 \\
Simple Average                & \textbf{80.34} & 61.29 & 88.63 & 69.53 & \textbf{71.61} & 64.12 & 75.01 & 67.66 & 69.40 & 76.99 & 74.27 & 73.00 \\
Confidence Weighted           & 78.44 & 61.73 & 88.66 & 69.09 & 68.93 & 63.51 & 74.21 & 66.11 & 65.51 & 77.78 & 73.33 & 71.12 \\
Majority Vote                 & 70.82 & 69.07 & 90.16 & \textbf{69.94} & 59.86 & 74.43 & 77.85 & 66.36 & 53.92 & \textbf{79.95} & 70.13 & 64.40 \\
Median Average                & 79.49 & 60.65 & 88.35 & 68.80 & 71.26 & 67.24 & 76.84 & \textbf{69.19} & 71.94 & 75.80 & 74.42 & 73.82 \\
Max Confidence                & 77.59 & 61.68 & 88.59 & 68.73 & 67.07 & 62.91 & 73.55 & 64.92 & 62.40 & 77.67 & 72.16 & 69.20 \\
Decision Fusion (Ours)             & \textbf{80.34} & 61.29 & 88.63 & 69.53 & \textbf{71.61} & 64.12 & 75.01 & 67.66 & 69.40 & 76.99 & 74.27 & 73.00 \\
Feature Fusion (Ours)               & 65.54 & \textbf{73.99} & \textbf{90.71} & 69.51 & 60.46 & 75.70 & \textbf{78.49} & 67.23 & \textbf{77.85} & 77.93 & \textbf{77.85} & \textbf{77.89} \\
Adaptive Fusion (Ours)              & \textbf{80.34} & 60.22 & 88.25 & 68.84 & 70.95 & 63.43 & 74.47 & 66.98 & 68.27 & 77.25 & 74.02 & 72.48 \\
\bottomrule
\end{tabular}
}
\end{table}

\begin{figure}[!htbp]
\centering
\includegraphics[width=0.48\linewidth]{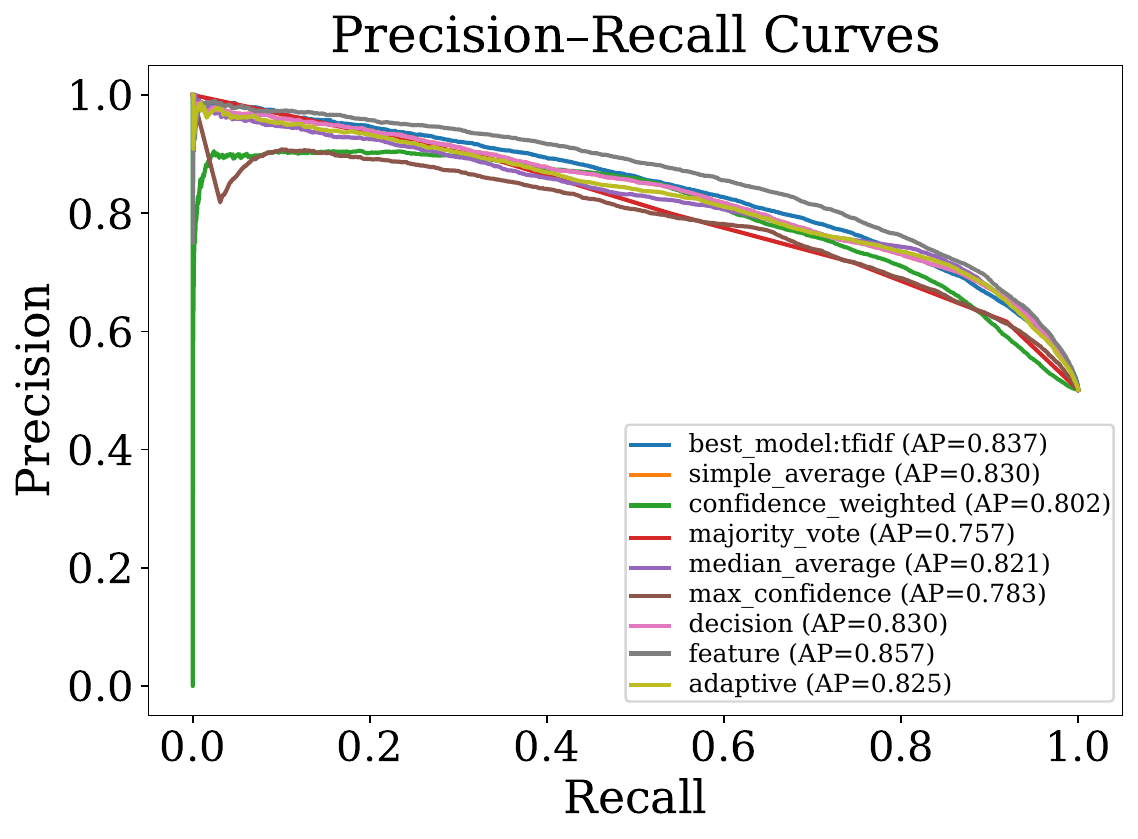}
\includegraphics[width=0.48\linewidth]{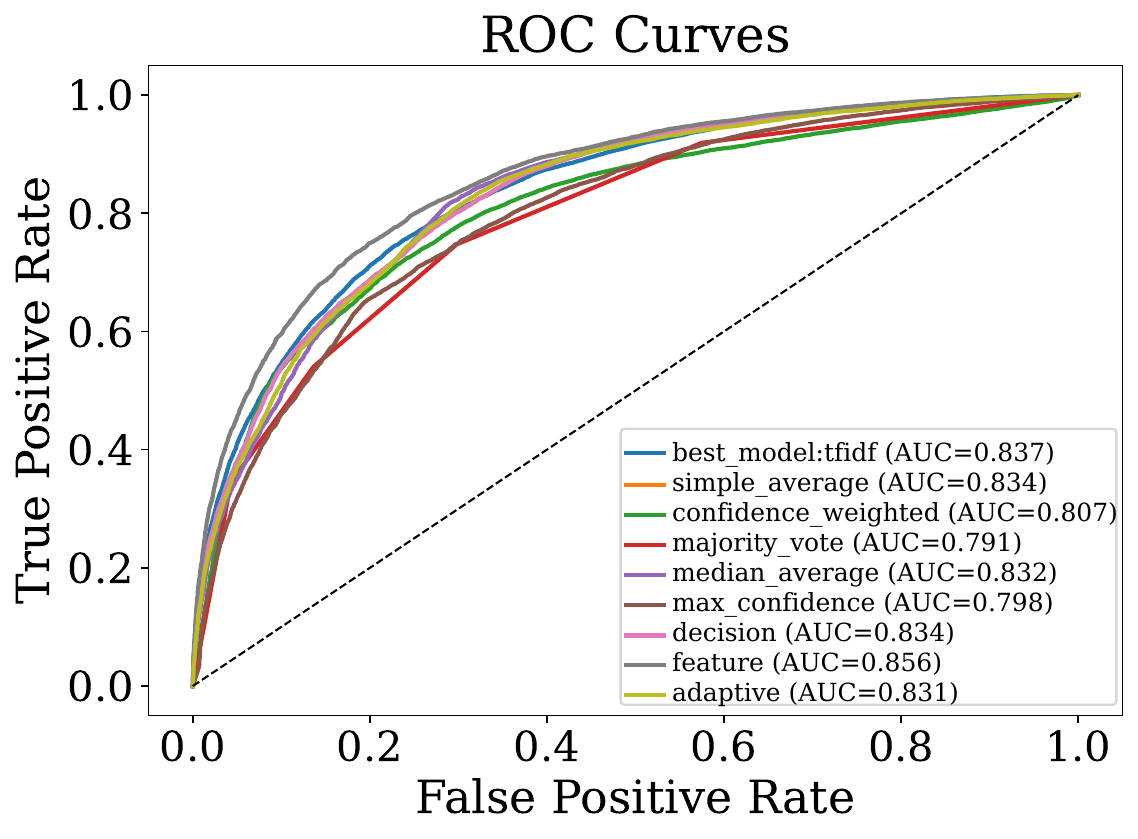}
\caption{PR curves and ROC curves on the Sentiment140 dataset.}
\label{fig:rq1_curves}
\end{figure}

On \textbf{Crowdflower}, attains the best Accuracy (90.71), with high Precision (73.99). Majority vote yields the best F1 (69.94), slightly above feature (69.51) and decision/simple (69.53). On \textbf{GoEmotions}, feature fusion gives the best Accuracy (78.49). Interestingly, median averaging yields the best F1 (69.19), exceeding decision/simple (67.66) and feature (67.23). On the large-scale \textbf{Sentiment140}, feature fusion is best on both Accuracy (77.85) and F1 (77.89), outperforming the best individual (76.11/76.43). Across datasets, feature-level fusion is the most reliable overall winner (best Acc on all three; best F1 on Sentiment140). We also present ROC and PR curves in Figure \ref{fig:rq1_curves} specifically for the Sentiment140 dataset due to its large scale (1.6M samples) and its balanced class distribution.

Feature-level fusion achieves the strongest performance with ROC-AUC of 0.856 and PR-AUC of 0.857, outperforming the best individual model (AUC 0.837, PR-AUC 0.837). Decision fusion and simple averaging follow closely (0.834 – 0.830), while majority vote and max-confidence lag behind (AUC below 0.800, PR-AUC as low as 0.757). These results confirm that structured fusion yields superior discriminative ability and more reliable precision–recall trade-offs than naive ensemble rules.

\subsection{Research Question 2}
\begin{figure}[!htbp]
    \centering
    \includegraphics[width=\textwidth]{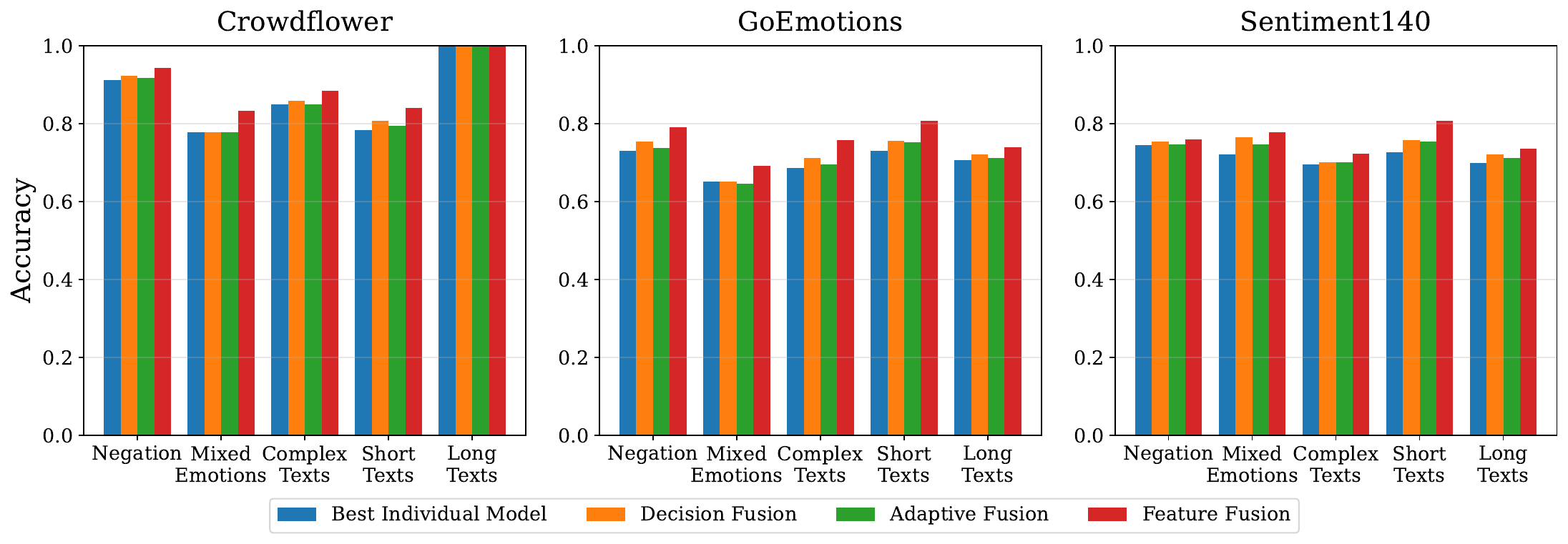}
    \caption{Accuracy of proposed framework and best individual model across datasets.}
    \label{fig:rq2_characteristics}
\end{figure}

The second research question, "How does fusion effectiveness vary across text characteristics?" examines whether different fusion strategies perform better on specific types of texts. Since models have complementary strengths, such as lexicon approaches excel on short texts while transformers capture complex semantics, we hypothesize that structured fusions can adapt to text conditions more effectively than either naive rules or single models. We then compare the best individual model, decision fusion, adaptive fusion, and feature fusion within each category. Figure~\ref{fig:rq2_characteristics} reports accuracy across the three datasets.

On \textbf{Crowdflower}, feature fusion consistently outperforms the other approaches across nearly all categories. It reaches about 0.95 accuracy on negation and nearly 1.0 on long texts, demonstrating that combining models captures subtle polarity shifts and benefits from richer context. Even on short texts, where lexicon-based methods usually excel, feature fusion attains 0.82, exceeding decision (0.75), adaptive (0.76), and the best individual model (0.75). This shows that feature-level integration can compensate for sparse signals by pooling complementary features. For \textbf{GoEmotions}, the advantage of feature fusion is again visible, particularly in challenging categories such as negation (0.77) and complex texts (0.73). Short texts also favor feature fusion (0.79 vs. 0.77 for decision and 0.73 for the best individual). Interestingly, on mixed emotions, the gap between strategies narrows, suggesting that all models find this phenomenon inherently difficult, and fusion only partially mitigates the challenge. On long texts, all methods converge near 1.0, reflecting that with sufficient context, both individual models and ensembles perform robustly. On the large-scale \textbf{Sentiment140}, feature fusion maintains a consistent edge across categories, but the margins are smaller than in the other datasets. It scores 0.74 on short texts versus 0.73 for the best individual, and 0.69 on complex texts versus 0.68 for decision and adaptive fusion. This suggests that on very large datasets, strong individual models already capture much of the available signal, and fusion strategies yield incremental but reliable improvements.

Overall, RQ2 confirms that the effectiveness of fusion varies with text characteristics. Feature fusion is most effective on negation, complex sentiment, and short texts, where single models struggle. Decision and adaptive fusion provide stable performance, often close to feature fusion, but without consistently surpassing it. On long texts, all approaches converge, highlighting that fusion adds the most value when input is limited or ambiguous.

\subsection{Research Question 3}
The third research question, "\textit{\textbf{Does the framework generalize across different model pools?}}", examines whether the proposed framework remains effective when applied to different sets of heterogeneous models. Across the three datasets, our results show that systematic fusion consistently matches or outperforms naive ensembles, which is illustrated in Table \ref{tabA:combo2} and Table \ref{tabA:combo3}. 

\begin{table}[!htbp]
\centering
\caption{Performance (percentage) of fusion strategies with combination of TextBlob + RoBERTa + SVM across datasets.}
\label{tabA:combo2}
\resizebox{\textwidth}{!}{%
\begin{tabular}{l|cccc|cccc|cccc}
\toprule
\multirow{2}{*}{Strategy} 
& \multicolumn{4}{c|}{Crowdflower} 
& \multicolumn{4}{c|}{GoEmotions} 
& \multicolumn{4}{c}{Sentiment140} \\
\cmidrule(lr){2-5} \cmidrule(lr){6-9} \cmidrule(lr){10-13}
& Recall & Prec & Acc & F1 
& Recall & Prec & Acc & F1 
& Recall & Prec & Acc & F1 \\
\midrule
Simple Average       & 58.33 & 73.25 & 78.11 & 64.68 & 62.92 & 71.06 & 75.90 & \textbf{66.70} & 52.02 & 75.58 & 64.17 & 56.33 \\
Confidence Weighted  & 59.33 & 72.69 & 78.11 & 64.92 & 62.05 & 71.53 & 75.94 & 66.39 & 49.61 & 75.61 & 63.65 & 54.98 \\
Majority Vote        & \textbf{62.83} & 72.65 & \textbf{80.43} & \textbf{66.91} & 59.84 & \textbf{72.35} & 75.97 & 65.53 & 49.34 & 76.06 & 63.60 & 54.95 \\
Median Average       & 60.67 & 73.09 & 79.18 & 66.30 & 61.52 & 71.71 & 75.98 & 66.25 & \textbf{52.14} & 75.55 & \textbf{64.37} & \textbf{56.40} \\
Max Confidence       & 60.17 & 72.53 & 78.93 & 65.68 & 59.73 & 71.95 & 75.60 & 65.09 & 45.86 & \textbf{76.24} & 62.61 & 52.07 \\
Decision Fusion (Ours) & 57.00 & 74.18 & 77.89 & 64.27 & \textbf{62.65} & 71.37 & 75.75 & 66.26 & 51.63 & 75.11 & 63.56 & 56.12 \\
Feature Fusion (Ours)  & 61.67 & 73.67 & 79.80 & 66.61 & 61.92 & 71.56 & \textbf{76.09} & 66.18 & 51.49 & 75.68 & 63.63 & 55.93 \\
Adaptive Fusion (Ours) & 55.50 & \textbf{75.83} & 77.39 & 64.13 & 62.06 & 71.99 & 75.96 & 66.16 & 50.39 & 75.46 & 63.58 & 55.79 \\
\bottomrule
\end{tabular}}
\end{table}

\begin{table}[!htbp]
\centering
\caption{Performance (percentage) of fusion strategies with combination of AFINN + BERT + XGBoost across datasets.}
\label{tabA:combo3}
\resizebox{\textwidth}{!}{%
\begin{tabular}{l|cccc|cccc|cccc}
\toprule
\multirow{2}{*}{Strategy} 
& \multicolumn{4}{c|}{Crowdflower} 
& \multicolumn{4}{c|}{GoEmotions} 
& \multicolumn{4}{c}{Sentiment140} \\
\cmidrule(lr){2-5} \cmidrule(lr){6-9} \cmidrule(lr){10-13}
& Recall & Prec & Acc & F1 
& Recall & Prec & Acc & F1 
& Recall & Prec & Acc & F1 \\
\midrule
Simple Average       & 75.66 & 61.94 & 88.16 & 68.20 & 71.08 & 64.37 & 75.20 & 67.29 & 66.78 & 76.07 & 73.63 & 71.59 \\
Confidence Weighted  & 75.16 & 62.19 & 88.14 & 68.20 & 70.09 & 64.99 & 75.18 & 67.23 & 63.23 & 76.40 & 72.46 & 70.07 \\
Majority Vote        & 71.00 & 67.62 & 89.91 & 69.02 & 62.59 & 72.98 & 77.83 & 66.59 & 56.29 & \textbf{78.17} & 70.22 & 65.34 \\
Median Average       & 74.00 & 62.38 & 87.89 & 67.80 & 70.83 & 66.18 & 76.36 & 67.27 & 69.85 & 75.71 & 74.71 & 72.70 \\
Max Confidence       & 73.66 & 62.47 & 88.20 & 67.90 & 69.05 & 64.91 & 75.14 & 66.60 & 60.55 & 76.89 & 72.06 & 68.38 \\
Decision Fusion (Ours) & 75.66 & 61.94 & 88.16 & 68.20 & 71.07 & 64.37 & 75.20 & 67.29 & 66.78 & 76.07 & 73.63 & 71.59 \\
Feature Fusion (Ours)  & 63.83 & \textbf{72.69} & \textbf{90.39} & \textbf{69.08} & 61.10 & \textbf{75.97} & \textbf{78.59} & 67.39 & \textbf{77.07} & 76.89 & \textbf{76.89} & \textbf{76.90} \\
Adaptive Fusion (Ours) & \textbf{79.16} & 60.62 & 88.27 & 68.51 & \textbf{72.52} & 63.67 & 75.18 & \textbf{67.82} & 66.44 & 76.89 & 73.49 & 71.36 \\
\bottomrule
\end{tabular}}
\end{table}

On \textbf{TextBlob + RoBERTa + SVM (Combo 2)}, naive ensembles produce mixed outcomes depending on the dataset. Majority voting achieves the highest accuracy on Crowdflower (80.4\%) and F1 (66.9\%), while median averaging performs best on GoEmotions (F1 66.3\%). These outcomes highlight that naive ensembles are sensitive to dataset conditions and fail to generalize consistently. In contrast, the SentiFuse strategies yield more stable performance. Feature fusion achieves competitive results across datasets (79.8\% accuracy and 66.6\% F1 on Crowdflower), while adaptive fusion reaches the highest precision on Crowdflower (75.8\%), indicating robustness in skewed distributions. Decision fusion remains reliable, closely tracking simple averaging but with lower variance. On \textbf{AFINN + BERT + XGBoost (Combo 3)}, structured fusion demonstrates clearer advantages. Feature fusion delivers the strongest overall balance, including 90.4\% accuracy on Crowdflower and 75.97\% precision with 78.59\% accuracy on GoEmotions, outperforming both naive and individual models. Adaptive fusion emphasizes recall, reaching 79.2\% on Crowdflower, while decision fusion again shows stable behavior aligned with simple averaging. On Sentiment140, feature fusion achieves the highest overall performance (76.9\% accuracy and F1), demonstrating that structured integration maintains benefits even in large-scale settings.  

Overall, RQ3 shows that SentiFuse generalizes effectively across heterogeneous model fusions. While naive ensembles occasionally perform well in isolated cases, their outcomes vary substantially across datasets. Structured methods, by contrast, consistently enhance reliability: feature fusion excels in fine-grained scenarios, adaptive fusion strengthens recall and robustness, and decision fusion provides stable performance. These results highlight that the framework is not tied to a particular model set but provides a general architecture for integrating diverse sentiment classifiers.  

\section{Conclusion}
In this paper, we introduced SentiFuse, a flexible and model-agnostic framework for sentiment analysis that integrates diverse models through a unified standardization and fusion pipeline. By supporting decision-level, feature-level, and adaptive fusion strategies, the framework improves performance across multiple datasets and challenging text phenomena such as negation, mixed emotions, and short or complex expressions. Our experiments demonstrate that no single strategy dominates in all cases: feature-level fusion provides strong overall gains, while adaptive and decision-level methods offer robustness in heterogeneous contexts. Feature-level fusion is particularly strong on categories such as negation and mixed emotions, where complementary cues from lexicon and neural embeddings align. Adaptive fusion tends to generalize better across varied categories by weighting models dynamically. These findings underline the methodological contribution of offering a general fusion architecture that can be applied across model families, the analytical contribution of revealing when and why certain fusion strategies succeed on specific text types, and the practical contribution of showing that such strategies generalize effectively across different model pools. Together, these contributions provide a more reliable foundation for sentiment classification and point toward broader applications where robustness and complementarity are essential. While SentiFuse introduces some additional inference cost from running multiple models, it remains lightweight: feature fusion adds only linear concatenation overhead, and adaptive fusion requires a small meta-classifier. Although our accuracies fluctuate around 80\%, this is consistent with prior state-of-the-art sentiment work on noisy social media text. Tweets and Reddit posts often include sarcasm, slang, and mixed emotions, and annotator agreement itself is rarely above 85–90\%. We expect that stronger models such as LLaMA or Gemma, if included in the pool, would further improve individual baselines. However, our framework is model-agnostic: fusion still helps in cases where even large models misclassify ambiguous or sarcastic inputs. Thus, SentiFuse can complement LLMs rather than compete with them. Future work will explore more context-aware adaptive mechanisms and extend the framework to multilingual and domain-specific settings.

\section*{Funding}
This work was supported by NSF - USA CNS-2219614.

\end{document}